# In Rain or Shine: Understanding and Overcoming Dataset Bias for Improving Robustness Against Weather Corruptions for Autonomous Vehicles

Aboli Marathe[1,3], Rahee Walambe[2,3]*, Ketan Kotecha[2,3]

abolim@cs.cmu.edu, rahee.walambe@sitpune.edu.in, director@sitpune.edu.in

1 Machine Learning Department, Carnegie Mellon University, Pittsburgh, PA, USA.
2 Symbiosis Institute of Technology (SIT), Symbiosis International (Deemed University), Pune, India
3 Symbiosis Centre for Applied Artificial Intelligence (SCAAI), Symbiosis International (Deemed University), Pune, India
* Corresponding Author

## Abstract

Several popular computer vision (CV) datasets, specifically employed for Object Detection (OD) in autonomous driving tasks exhibit biases due to a range of factors including weather and lighting conditions. These biases may impair a model's generalizability, rendering it ineffective for OD in novel and unseen datasets. Especially, in autonomous driving, it may prove extremely high risk and unsafe for the vehicle and its surroundings. This work focuses on understanding these datasets better by identifying such "good-weather" bias. Methods to mitigate such bias which allow the OD models to perform better and improve robustness are also demonstrated. A simple yet effective OD framework for studying bias mitigation is proposed. Using this framework, the performance on popular datasets is analyzed and a significant difference in model performance is observed. Additionally, a knowledge transfer technique and a synthetic image corruption technique are proposed to mitigate the identified bias. Finally, using the DAWN dataset, the findings are validated on the OD task, demonstrating the effectiveness of our techniques in mitigating real-world "good-weather" bias. The experiments show that the proposed techniques outperform baseline methods by an average fourfold improvement.

## 1. Introduction

Rapid development of self-driving vehicles has sparked a surge of interest in building their software and hardware systems which enable them to carry out various tasks effectively. The goal of these systems is to ensure that the vehicle behaves consistently under all scenarios. The vehicles must be able to make ethical decisions for the safety of passengers and pedestrians. One major component of these multifaceted systems is the OD module which observes the scenes, converts them into useful information and detects objects. The ability of this module to function satisfactorily under adverse conditions relies on the robustness and precision of the supporting machine learning models. However, in machine learning tasks, the performance of models is highly dependent on the data it is fed with. If the training data contains bias, the model is bound to generate biased outcomes. The standard datasets used for training these OD modules typically have only the images captured in good weather conditions. When such data is used for OD module training, it does not perform well for images containing weather corruptions at the testing time. This can lead to accidents and even loss of life. Hence, it is essential that the OD module can examine their surroundings objectively in a variety of environmental circumstances in an unbiased manner.

Robust OD in out-of-distribution scenarios is one of the interesting challenges to autonomous driving that is gaining attention in recent years. As autonomous vehicles (AV) experience a wide variety of scenarios on real roads consisting of various environmental conditions, road conditions, types of objects, etc., it is necessary to train robust models which can effectively support the software driving the vehicles' decisions under this variability and heterogeneity of situation. Due to the diversity of scenes that AVs face, it becomes difficult to simulate samples for every possible situation and thus generic techniques of achieving robustness are sought. One out-of-distribution scenario that puzzles modern software is adverse weather conditions. These conditions can range from heavy snowfall, haze, smog, heavy fog, rainfall, to even dust tornados etc. Each of these conditions presents a unique hindrance to CV. in the challenge with such conditions is not only the lack of clarity in the image but blurred boundaries and added noise in the form of droplets, snowballs, or sleet. This noise obscures the images and confuses detection models that are accustomed to perfect weather scenarios. As several open datasets have been collected under ideal day scenarios (daytime, clear skies and absence of cloudy weather), we hypothesize that machine learning models that are trained on these massive datasets are directly biased towards "good-weather". This bias affects their ability to perform in adverse weather scenarios which presents a major risk to the safety of AVs and their surroundings subsequently leading to accidents and collisions and in some cases outright failure in OD. As these self-driving cars navigate through pedestrians, animals, and buildings using their OD-driven guidance systems, the inability to detect objects in adverse weather is a major security threat.

To that end, in this work, this primary challenge is considered, and it is proved that the benchmark datasets are biased towards good weather. Additionally, how to measure such 'good weather bias' and which methods to apply for such bias measurement and mitigation are also demonstrated.

Primary contributions of this work are as follows:
1. Identification of bias prevailing in several common benchmark datasets for OD in autonomous vehicles.
2. Measurement of the bias identified on the DAWN dataset using standard OD evaluation metrics.
3. Demonstration of novel methods for improving the robustness of general models in adverse weather-specific out-of-distribution scenarios using image alteration techniques.

The paper is organized as: Section 2 discusses the related work specific to OD and bias pertaining to autonomous driving. Section 3 introduces the target dataset and the proposed bias identification and mitigation framework. Section 4 consists of the experimentation and results obtained by the framework and a detailed discussion of the detection results. Section 5 concludes the work with potential future research directions.

**2. Related Work**

AV systems leverage several machine learning techniques for complex tasks like lane predictions, trajectory prediction, object (pedestrian and car) detection and tracking, speed estimation, decision making, and collision avoidance etc. However, the accurate and unbiased outcome of these tasks is primarily based on the training data. If the training data is biased towards a certain situation, the outcome of the tasks is also biased. This is very relevant for OD tasks, wherein, if the training data is not generic, the model does not generalize well to various environmental conditions in the real world. In this section, a few selected works, pertaining to OD and bias identification for autonomous driving systems are discussed.

    a. Object detection

The creation of large autonomous driving scene datasets in recent years has aided the development of AV systems and research. Several 2D RGB datasets with bounding box annotations are employed for detection tasks including

Cityscapes (Cordts et al. 2016), BDD100k (Yu et al. 2020), Apolloscape (Huang et al. 2018), KITTI Vision Benchmark Suite (Geiger et al. 2012) and CamVid (Brostow et al. 2008).

In 2D datasets, the task of 2D OD grew popular with the growing use of multi-stage and single-stage detectors like SSD, Faster R-CNN, and YOLO (Liu et al. 2016; Ren et al. 2015; Redmon et al. 2016). Taking 2D detectors forward, 3D detection expanded with Stereo RCNN (Li et al. 2019), AVOD (Ku et al. 2018), MVLidarNet (Chen et al. 2020) and MVF algorithm (Zhou et al. 2020) bringing new perspectives to the task of OD.

Adapting to out-of-distribution scenarios has become an essential feature of driving systems and thus several works have emerged to solve this problem recently. Novel planning methods (Filos et al. 2020), networks for identifying objects from unknown classes (Wong et al. 2020) and multi-modal fusion transformer methods (Prakash et al. 2021) have made progress in identifying anomalous features and show robust performance under diverse conditions. Specific challenges like multiscale OD (Walambe et al. 2021), pedestrian detection in crowds (Marathe et al. 2021) and detection under adverse weather (Walambe et al. 2021) have also been tackled using ensemble methods and data augmentation.

b. Bias

One of the primary challenges posed to the implementation of ethical AI is bias. Bias is generally introduced in AI systems through data, people and model design. Data bias has been the reason for various issues with the implementation of ethical and fair AI (Torralba and Efros 2011; Khosla et al. 2013). It generates outcomes that are discriminatory and affect the opportunities and resource allocation for under-represented demographic groups. However, when it comes to high-risk applications such as autonomous driving, it is not just about fairness and ethics but can pose an even greater challenge specific to the safety of the surroundings and the AV itself.

As AVs deal with changing driving conditions and unpredictable scenarios, there is scope for many forms of bias to creep its way into the model predictions. Specifically, if the training data is only specific to certain conditions, then the model is incapable of making accurate decisions in the real world. In particular, several unique forms of bias have been identified in autonomous driving systems by previous works and novel methods to mitigate them were proposed as well (Jo et al. 2013; Bando et al. 2011; Danks and London 2017). One form of bias identified was the GPS error bias and was proposed to be corrected using additional information sources like camera vision systems (Jo et al. 2013). Kalman filters were shown to have good results in estimating sensor bias in AVs, which is useful for estimating altitude correctly (Bando et al. 2011). Multiple possible sources of algorithmic bias were discussed from the perspective of autonomous systems, for e.g., the performance of a model trained on data from a certain city when used in another city (Danks and London 2017).

However, there is no study that specifically considers these benchmark datasets for OD to understand the bias in them. In this work, we propose the framework to identify, measure and mitigate such bias in these benchmark datasets.

## 3. Methodology

A. Bias Identification and Mitigation Framework

We use a SSD model trained with 512x512 images on multiple datasets with MobileNet V1 as the base model to learn the object classes and form predictions on the target dataset (Liu et al. 2016). The same loss function was used across all implemented SSD models. The loss function (Equation 1) is the weighted sum of 2 loss functions, the confidence loss and the localization loss.

$$L(x,c,l,g) = \frac{1}{N}\Big(L_{conf}(x,c) + \alpha L_{loc}(x,l,g)\Big) \qquad [1]$$

where N is the number of matched default boxes, and the localization loss is the Smooth L1 loss (Girshick 2015) between the predicted box (l) and the ground truth box (g) parameters. We used cross-entropy loss (Eq 2) as confidence loss.

$$CE = -\sum_{c=1}^{M} y_{o,c} \log(p_{o,c}) \qquad [2]$$

Where M - number of classes (dog, cat, fish), log - the natural log, y - binary indicator (0 or 1) if class label c is the correct classification for observation o , p - predicted probability observation o is of class c.

For the second loss smoothL1 (Eq 3) loss is used, the absolute value of the difference between the prediction and the ground-truth.

$$loss(x,y) = \begin{array}{ll} 0.5(x-y)^2 & if\ |x-y| < 1; \\ |x-y| - 0.5 & otherwise \end{array} \qquad [3]$$

The goal of the study is to analyze the relationship between object classes detected by the models through the framework. Variations of the MobileNet SSD model were trained on different datasets and the detection results are obtained as per the procedure shown in Fig. 2. We attempt to piece together more loose relationships between clean images, corrupted images and detected classes for learning which methods of image modification yield enhanced robustness.

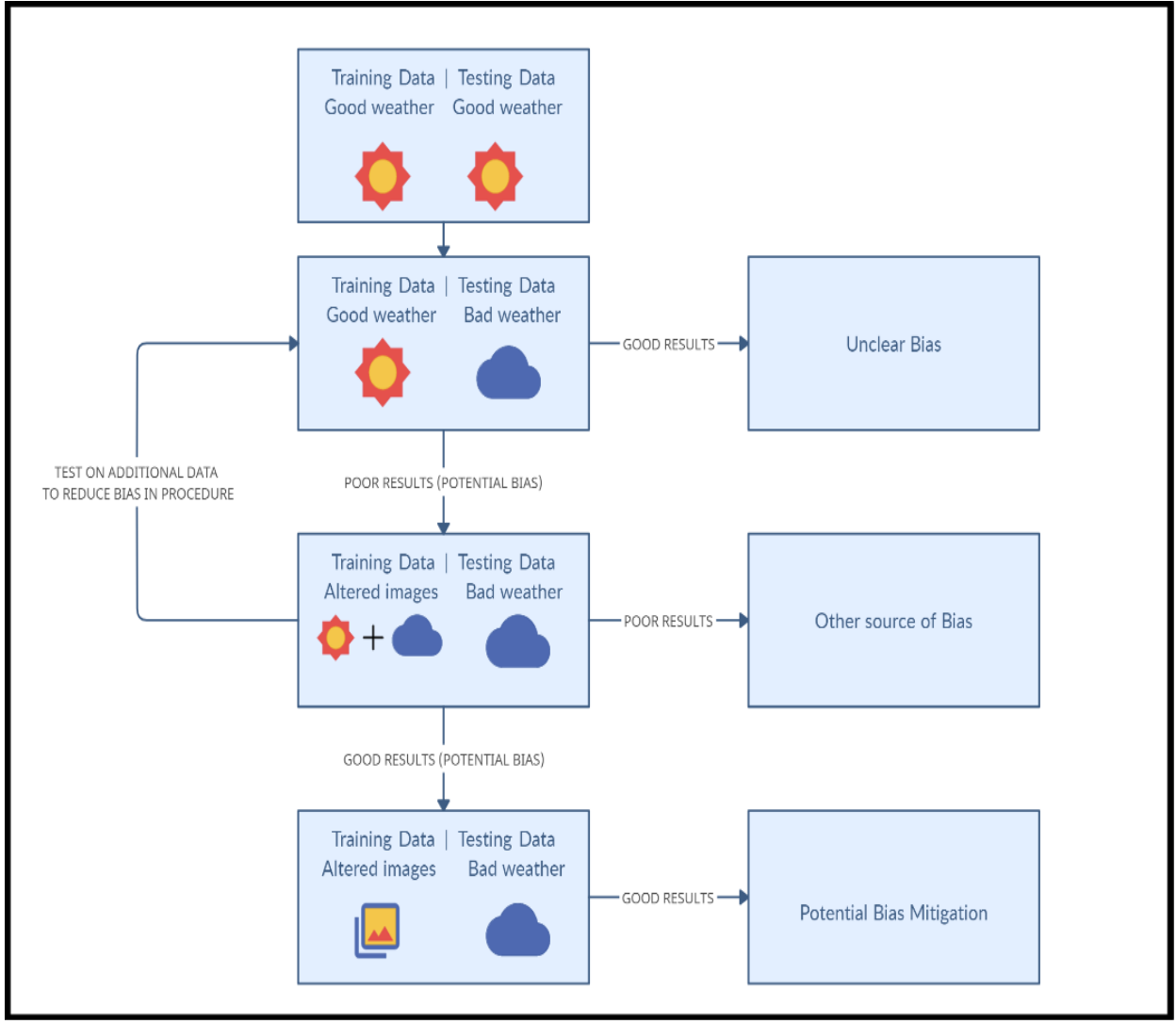

**Fig. 1:** Bias Identification and Mitigation Framework

As shown in the flowchart in Fig. 3, there are 5 steps involved in the proposed bias identification and mitigation framework.

1. The first step consisted of training the baseline model on a perfect weather (uncorrupted) dataset. Then the model performance was examined on a test perfect weather (uncorrupted) dataset. The OD results of common object classes using evaluation metrics are obtained. The purpose of this step is to establish baseline performance of the model on a perfect dataset captured under perfect weather conditions in order to determine if the model can satisfactorily identify objects and learn features from input data.
2. The second step consisted of training the baseline model on a perfect weather(uncorrupted) dataset. Then the model performance was examined on a test adverse weather (corrupted) dataset. The OD results of common object classes using evaluation metrics are obtained. The purpose of this step is to find differing results from Step 1 in order to determine if the model can perform robustly under these 2 conditions. If the model performs perfectly, then there is no need for taking measures to increase robustness. However, if the model fails in OD, it must be investigated further, and the performance difference must be measured.
3. The third step consisted of training the baseline model on a mixture of perfect weather(uncorrupted) and adverse weather(corrupted) images. The model performance is then evaluated on a test adverse weather (corrupted) dataset and OD results of common object classes are evaluated using evaluation metrics. The purpose of this step is to find differing results from Step 2 in order to determine if the model is able to perform robustly under these 2 conditions. If the model fails in OD, we must investigate further and measure the performance difference.
4. In Step 1-3, a measurable bias in OD is identified using the mAP and class wise AP while testing models on DAWN dataset. However, to prevent another form of bias from creeping into this experiment due to the use of only one perfect weather dataset, we carry out the same experimentation using another distinct dataset.
5. Bias Mitigation: After confirming the bias in detection, several strategies are proposed to mitigate the bias for creating robust models.

    a. Knowledge transfer: In this method, transfer learning between the good and bad weather dataset is implemented for the model to learn representations effectively and differentiate between object classes.
    b. Synthetic Image Corruption: In this method, the images from training dataset are purposely corrupted for the model to learn representation robustly without target data.

The performance after applying these techniques is measured and the overall results are compared to validate the procedure and proposed techniques.

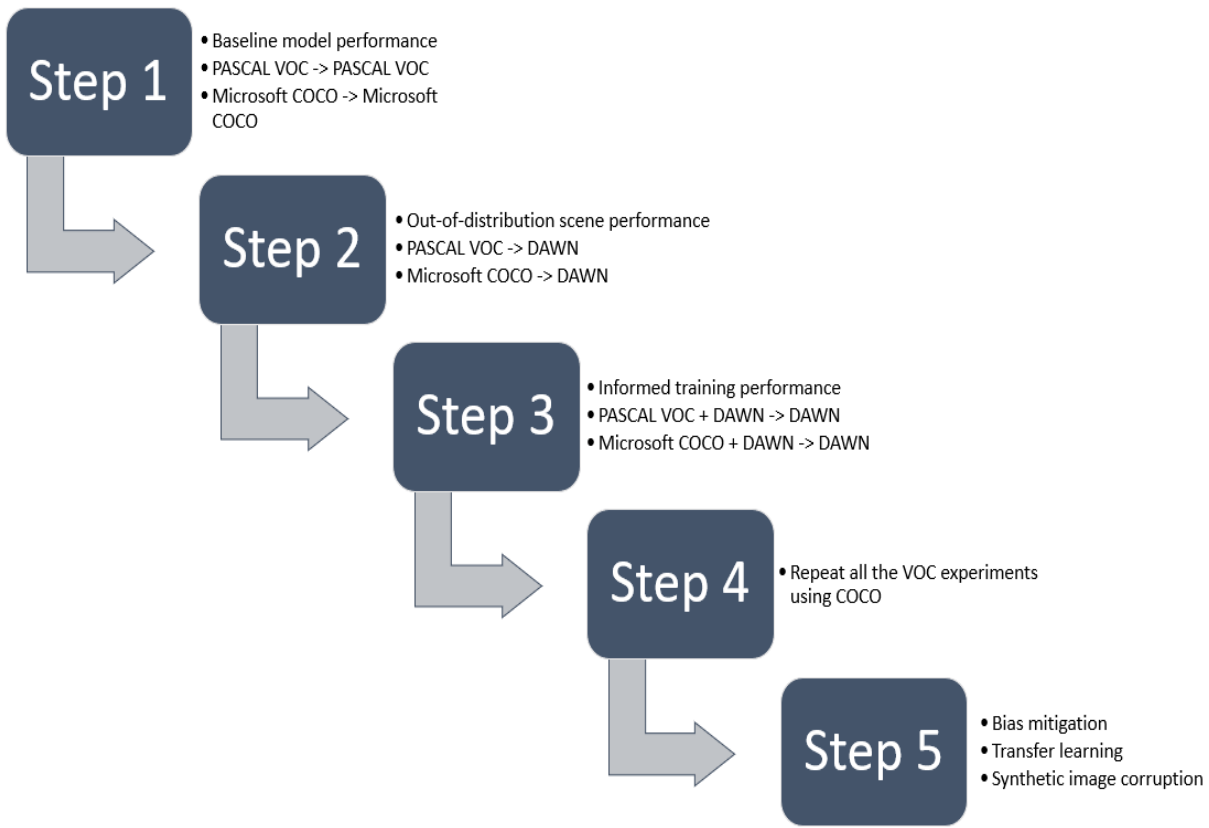

**Fig. 2:** Bias M & I Procedure followed in the study

B. Dataset Description

As seen in the proposed framework, we require at least one adverse weather condition dataset and two good weather datasets for testing the results and establishing the baseline. For this testing purpose we utilized the DAWN dataset (Kenk and Hassaballah 2020) as it captures traffic scenes and poor weather conditions. For the baseline determination, we used the Microsoft COCO (Lin et al. 2014) and PASCAL VOC datasets (Everingham et al. 2007; Everingham et al. 2012) which are a few popular visual recognition benchmarks with ideal weather images. A stark difference in image clarity can be seen in these datasets, as observable in Figure 3.

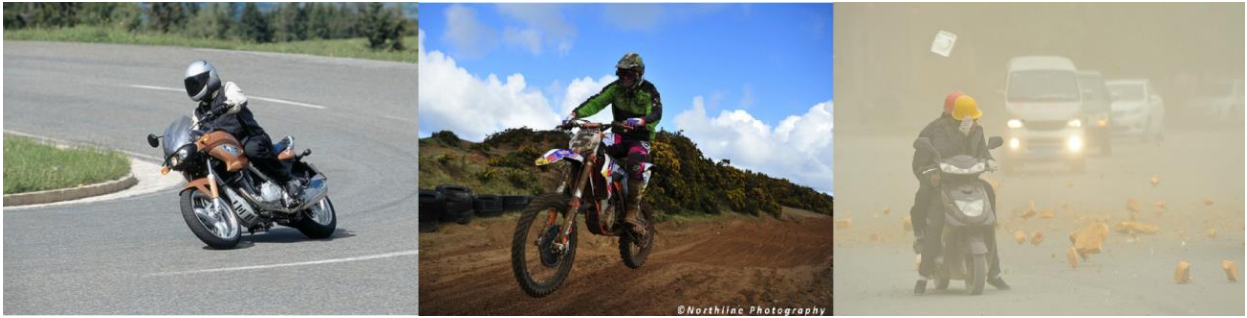

**Fig. 3:** Samples from PASCAL VOC 2012, Microsoft COCO and DAWN datasets.

The DAWN dataset is a large vehicle detection dataset which has captured images of driving scenes in adverse weather conditions (Kenk and Hassaballah 2020). It consists of 1000 images from real-traffic scenes as seen in 4 adverse weather conditions: fog, snow, rain, and sandstorms (Refer Fig 1). The dataset contains significant diversity of vehicle category, size, orientation, pose, illumination, position, and occlusion thus making it a robust choice for bias identification. The images given in the dataset are of size 1,280 × 856 pixels. The images have been annotated with 2D annotations(boxes) with 6 object classes namely car, bus, truck, motorcycle, person and bicycle.

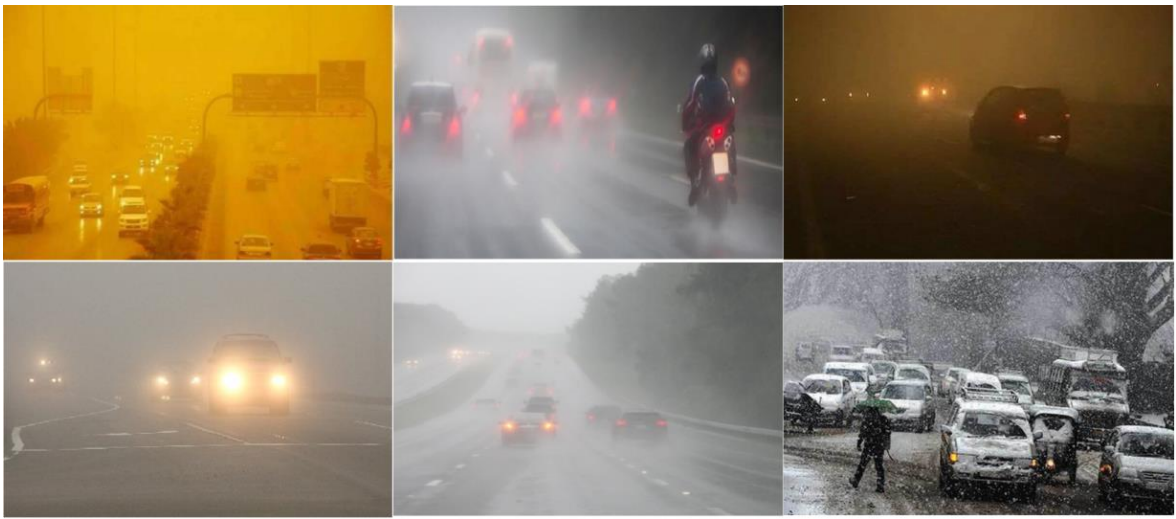

**Fig. 4:** Weather conditions in DAWN dataset from top left corner towards right a) Dust tornado b) fog c) haze d) mist e) rain f) snowstorm.

# 4. Experiments and Results

   a. Bias Identification

Three SSD Mobilenet V1 models are iteratively implemented and trained on increasingly corrupted training data to study the effect of training image quality on precision of ODs. For all models, an SGD optimizer with a learning rate of $1 \times 10^{-3}$ and a batch size 16 is used. Various experiments are carried out by tuning the hyperparameters namely, learning rate and the batch size. For identifying potential bias an investigative procedure is followed and results are reported after the carrying out the following steps:

**Implementation of Stage 1:** For the baseline determination, subsets of mixed Pascal 2007 (Everingham et al. 2007) and Pascal VOC 2012 dataset (Everingham et al. 2012) are used for training and testing the SSD model. The training image set is the union of 2007trainval and 2012trainval and the validation image set is 2007test. This subset formed a collection of 16551 images for training and 4952 images for testing. As Pascal VOC supports 20 classes, however only 4 relevant classes to the DAWN dataset were selected for the analysis namely, car, bus, person and bicycle. The SSD model trained on uncorrupted data performs very well on uncorrupted data, as expected and reaches over 70 AP on all classes with 74.95 mAP as shown in Table 1. Therefore, the baseline was established, and satisfactory results were obtained for OD with the MobileNet SSD model.

**Implementation of Stage 2:** For measuring the potential bias, the model was trained on the same training data as Step 1 and tested the performance on the 1000 adverse weather condition images of the DAWN dataset (Kenk and Hassaballah 2020). The model fails to perform satisfactorily on the corrupted images and reaches 3.75 mAP with nearly 7 AP on most classes as shown in Table 1. It fails to detect buses and bicycles completely and shows low performance across all classes. Despite the similarity in object classes, the inability of the model to detect objects in the DAWN dataset leads to potential bias hindering the OD.

**Implementation of Stage 3:** Finally, the model was trained on a mixture of bad and good weather images from PASCAL VOC 2012, 2007 and DAWN data and tested the performance on the 1000 adverse weather condition images of the DAWN dataset (Kenk and Hassaballah 2020). The model fails to perform satisfactorily on the corrupted images and reaches 5.5 mAP with nearly 7 AP on most classes as seen in Table 1. It fails to detect bicycles completely but shows improved performance (7 AP) on the bus class. Mixing a small percentage of images is able to counter the bias minutely, however this method is insufficient for boosting overall performance.

**Implementation of Stage 4-5**: In Step 1-3, we established that there was a measurable bias in OD while testing models on DAWN dataset. However, to prevent another form of bias from creeping into this experiment due to the use of only one perfect weather dataset (Pascal VOC), the same experimentation was carried out using the Microsoft COCO dataset (Lin et al. 2014).

   c. For the baseline determination, subsets of Microsoft COCO 2017 dataset (Lin et al. 2014) are used. The training imageset is train2017 and validation imageset is val2017. This subset formed a collection of 118000 images for training and 5000 images for testing. As COCO supports 80 classes, however only 4 relevant classes to the DAWN dataset were selected for the analysis namely, car, bus, person and bicycle. The SSD model trained on uncorrupted data performs very well on uncorrupted data, as expected and reaches over 47 AP on bus class with 28.37 mAP as shown in Table 1. Therefore, the baseline was established, and satisfactory OD results were obtained with the MobileNet SSD model.

d. In measuring the potential bias, the model was trained on the same training data as Step 4-part a and tested the performance on the 1000 adverse weather condition images of the DAWN dataset (Kenk and Hassaballah 2020). The model fails to perform satisfactorily on the corrupted images and reaches 9.5 mAP with 29 AP on person classes as shown in Table 1. It fails to detect bicycles completely and shows low performance across all classes. Despite the similarity in object classes, the inability of the model to detect objects in the DAWN dataset leads to potential bias hindering the OD.

e. Finally, the model was trained on a mixture of bad and good weather images from Microsoft COCO 2017 and DAWN data and tested the performance on the 1000 adverse weather condition images of the DAWN dataset (Kenk and Hassaballah 2020). The model fails to perform satisfactorily on the corrupted images and reaches 15.25 mAP with nearly 29 AP on person classes as shown in Table 1. It fails to detect bicycles completely but shows improved performance (26 AP) on the bus class. Mixing a small percentage of images can counter the bias minutely, however this method is insufficient for boosting overall performance.

Table 1: Performance on good and bad weather datasets. The models are compared as a part of stage 1, 2 and 3, bias identification.

| Model | Class 0 AP [car] | Class 1 AP [bus] | Class 3 AP [person] | Class 5 AP [bicycle] | MAP |
|---|---|---|---|---|---|
| **VOC** | | | | | |
| Stage 1 [Tested on VOC 2012] | 72.84 | 77.42 | 70.42 | 79.13 | 74.95 |
| Stage2 [Tested on DAWN] | 7 | 0.0 | 8 | 0.0 | 3.75 |
| Stage 3 [Tested on DAWN] | 7 | 7 | 8 | 0 | 5.5 |
| **COCO** | | | | | |
| Stage 1 [Tested on COCO 2017] | 18.80 | 47.1 | 32.5 | 15.10 | 28.37 |
| Stage 2 [Tested on DAWN] | 6 | 3 | 29 | 0.0 | 9.5 |
| Stage 3 [Tested on DAWN] | 6 | 26 | 29 | 0 | 15.25 |

b. **Bias Mitigation**

As observed in the above procedure, the models are clearly biased against adverse weather and are unable to perform satisfactorily. In this section, two methods of bias mitigation are presented. First, using images of the target dataset and the second in the absence of target images.

1. Knowledge Transfer:
   The model is first trained on the Pascal VOC 2017 dataset and then learned by fine tuning on the DAWN dataset. As displayed in Table 2, there is a rise by 1.75 mAP using this method as the mAP reaches 5.5 and the model can detect buses with increased 7 AP.
2. Synthetic image corruption:
   Here a Double Gaussian Blurring technique (Walambe et al. 2021) is employed to corrupt the images of the Pascal VOC 2017 dataset. The model is first trained on the corrupted Pascal VOC 2017 dataset and then tested its performance on the DAWN dataset. As seen in Table 2, there is a sharp rise by 22.5 mAP using this method as the mAP reaches 26.25 mAP and the model can detect bicycles with 92 AP.

Table 2: Performance of bias mitigation techniques on DAWN dataset (Kenk and Hassaballah 2020).

| Model | Class 0 AP [car] | Class 1 AP [bus] | Class 3 AP [person] | Class 5 AP [bicycle] | mAP |
|---|---|---|---|---|---|
| Original trained on VOC | 7 | 0.0 | 8 | 0.0 | 3.75 |
| Technique 1: Transfer Learning | 7 | 7 | 8 | 0 | 5.5 |
| Technique 2: Double Gaussian Blurring | 6 | 3 | 4 | 92 | 26.25 |

## 5. Conclusion

In this paper, we demonstrated the problem of object detection in adverse weather conditions by showing that standard classifiers perform poorly when anomalous weather corrupts their vision. We discovered significant "good-weather bias" in the models tested on DAWN dataset, thus opening the discussion for the impact of this bias on future work. This bias is observed in models that have never encountered noisy weather-corrupted data, thus limiting their OD capabilities. Finally, we proposed two techniques for mitigating this "good weather bias" in adverse weather conditions and conducted thorough experimentation to prove the same.

The first technique which proved effective was transfer learning, by using actual images of the target dataset to make the model more robust. This method increased the mAP by 1.75 and particularly improved the model's ability to detect buses in the corrupted images. However, this technique performs similar to the method of mixing images of good and bad weather for training. The second method we discovered was double Gaussian blurring corruption, by using images of PASCAL VOC 2017 baseline dataset to make the model more robust. This method boosted the mAP by 7 times and particularly improved the model's ability to detect bicycles in the corrupted images. As bicycles are shown to be difficult objects to detect in adverse weather through the experiments, this improvement bodes well for the robustness of this model.

This study explores how weather influences model considerations, opening up the opportunity to explore further challenges to robust OD in AVs. The factor of human trust and safety which supports the adoption of AVs is dependent on the elimination of bias and vulnerabilities from potential attacks. This problem in self-driving cars inspires the search for similar problems like object background bias, location bias, time-of-day bias, perspective bias and more. As several research works have already begun working in these directions, investigating potential biases in autonomous driving systems is gaining momentum. Through this work, we would like to encourage progress in this line of computer vision research, towards identifying and mitigating bias in the real-world.

## 6. Data Availability

All the datasets used in the current study are accessible and can be found at:
1. DAWN dataset (Kenk and Hassaballah 2020) available at DAWN Repository.
2. Microsoft COCO (Lin et al. 2014) available at COCO Repository.
3. PASCAL VOC datasets (Everingham et al. 2007; Everingham et al. 2012) available at PASCAL VOC Repository.